%% file: main.tex
\documentclass{article}

\usepackage{spconf}
\usepackage{amsmath}
\usepackage{graphicx}
\usepackage{algorithm,}
\usepackage{algorithmic}
\usepackage{amssymb}
\usepackage{booktabs}
\usepackage{multirow}
\usepackage{adjustbox}
\usepackage{amsfonts}
\usepackage{import}
\usepackage{pgfplots}
\usepackage{tikz}
\usepackage{hyperref}

\pgfplotsset{compat = newest}

\title{Domain Adapting Ability of Self-Supervised Learning for Face Recognition}
\name{Chun-Hsien Lin, Student Member, IEEE, and Bing-Fei Wu, Fellow, IEEE\thanks{This work was supported by the Ministry of Science Technology
under Grant MOST 108-2638-E-009-001-MY2.}}
\address{National Chiao Tung University\\ 
	Institute of Electrical and Control Engineering\\
	1001 University Road, Hsinchu, Taiwan 300, ROC}
\begin{document}
%
\maketitle
\begin{abstract}
Although deep convolutional networks have achieved great performance in face recognition tasks, the challenge of domain discrepancy still exists in real world applications. Lack of domain coverage of training data (source domain) makes the learned models degenerate in a testing scenario (target domain). In face recognition tasks, classes in two domains are usually different, so classical domain adaptation approaches, assuming there are shared classes in domains, may not be reasonable solutions for this problem. In this paper, self-supervised learning is adopted to learn a better embedding space where the subjects in target domain are more distinguishable. The learning goal is maximizing the similarity between the embeddings of each image and its mirror in both domains. The experiments show its competitive results compared with prior works. To know the reason why it can achieve such performance, we further discuss how this approach affects the learning of embeddings.
\end{abstract}

\section{Introduction}

With the growing of dataset and model capacity, the accuracy of face recognition is getting higher. Face recognition has been an efficient tool for authentication and has been widely deployed to many applications. Even if a learned model can perform nearly perfect in benchmark datasets, it may fail in some scenarios. The primary reason is the domain discrepancy between the training data (source domain) and testing data (target domain). The factors causing the domain discrepancy may be illumination, blur, pose, race, gender, or age. Most training images are collected on Internet, which makes the training set contain various domain information. However, only few samples come from same domain, so it is hard to train a model with good generalization. A straight forward solution is to fine-tune the learned model on target scenarios with labels. This approach is not practice, however, because labeling enough data is too expensive. Therefore, how to leverage the unlabeled data in target domains becomes an important issue, and this is what domain adaptation does.

Current approaches of domain adaptation \cite{wang2018deep} can be divided into two main branches: adapting classifier and adapting embedding space. The former is tunning the classifiers to adapt to the distribution in target domains. The later is tunning the embedding functions (feature extractors) to find a common embedding space where the distributions of tow domains are aligned. As for the embedding distribution of source domain, areas near the cluster centers are the high density places. Aligning the distributions of two domains is similar to uniformly assigning the embeddings of target domain to the classes of source domain, so both approaches are relied on an assumption that source and target domains are shared classes. Oppositely, in face recognition tasks, the classes are different in domains. It is not reasonable to apply the approaches above to solve this unique problem. Despite the different preconditions, the techniques of embedding alignment is still widely adopted in most existing works, \cite{wang2020deep}\cite{wang2019racial}\cite{luo2018deep}\cite{sohn2017unsupervised}\cite{hong2017sspp}, to mitigate the domain shift of face recognition.

In this paper, to avoid using embedding alignment, we use self-supervised learning to maximize the self-similarity of each sample. In this way, the embedding distribution of target domain is not aligned to the clusters of source domain, but organizes itself. Tested in IJB-A \cite{klare2015pushing}, IJB-B \cite{whitelam2017iarpa}, and IJB-C \cite{maze2018iarpa} datasets, the proposed approach shows its ability of domain transfer. It is interesting that we find the improvement is achieve by lowering inter-class similarity rather than enlarging intra-class similarity. Unlike prior works, \cite{wang2020deep}\cite{wang2019racial}\cite{arachchilage2020ssdl}, using larger models, like ResNet \cite{he2016deep} or VGG \cite{BMVC2015_41}, we use MobileFaceNet \cite{chen2018mobilefacenets} as the backbone. In spite of smaller capacity, the testing results in experiments are competitive and even better. The contributions of this research can be summarized into two parts. First, we propose a novel method to perform domain transfer for face recognition. Second, we further analyze the mechanism beyond the learning algorithm. 


\section{Related Work}

\subsection{Domain adaptation for face recognition}

Because of the unique issues of face recognition, few researches focus on domain adaptation for face recognition. Some prior works, \cite{luo2018deep}\cite{sohn2017unsupervised}\cite{hong2017sspp}, adopt embedding alignment methods to be the core algorithm directly, like Maximum Mean Discrepancy (MMD), \cite{long2018transferable}, or adversarial learning, \cite{ganin2015unsupervised}, which does not really reduce the domain shift existing in face recognition. To compensate embedding alignment, pseudo labels generated by clustering on target domain are utilized to train another cluster distribution are proposed in \cite{wang2020deep} and \cite{wang2019racial}. Arachchilage et al. \cite{arachchilage2020ssdl} focus on the task of video frames. With the identity consistency in a video, clustering can be more accurate. After clustering, triplets are mined to fine-tune the model with a modified triplet loss. However, clustering relies on prior knowledges of class number or cluster margin, which may be a non-trivial work when setting these parameters.


\subsection{Self-supervised learning}

To leverage unlabeled data, some auxiliary tasks are designed in training losses based on the prior knowledges of data. For vision tasks, the simplest task is self-similarity. Chen et al. \cite{chen2020simple} augment an image, and adopt contrastive learning to minimizing their distances while maximizing the distances among different images in the embedding space. The method in \cite{chen2020exploring} achieves better performance by adopting cosine similarity with stop-gradient operation. Its transfer ability is briefly illustrated, but it is limited. Therefore, we mainly refer this research to design an adapting loss for face recognition.  



\section{Proposed Approach}

There are two datasets in our case. The source dataset (training dataset) is denoted as $X^s=\{x^s_i,y^s_i\}^N_{i=1}$, where $x^s_i$ is a facial image in source domain, $y^s_i$ is one-hot encoding label for $x^s_i$, and $N$ is the number of source images. The target dataset is denoted as $X^t=\{x^s_j\}^M_{j=1}$, where $x^t_j$ is a facial image in target domain, and $M$ is the number of target images. For convenience, an arbitrary image, a source image and a target image are denoted as $x$, $x^s$ and $x^t$ respectively. 

\subsection{Learning embeddings}

High recognition accuracy replies on a good embedding function. The purpose of an embedding function $f(x)$ is mapping images on a lower dimensional space where embeddings from same classes are closer while embeddings from different classes are farther. We can train an embedding function with an extra classifier, $\hat{y}(f(x))$, by cross entropy, which is simple and robust. The cross entropy loss is defined as:
\begin{equation}
L_c=-\sum_{i=1}^{N}y^s_i\log{\hat{y}(f(x^s_i))}
\end{equation}
Since training dataset is usually large, to accelerate the convergence, focal loss \cite{lin2017focal} is adopted to encourage the learning of hard samples, so the loss can be modified as:
\begin{equation}
L_c = -\sum_{i=1}^{N}(1-\hat{y}(f(x^s_i)))^{\gamma}y^s_i\log{\hat{y}(f(x^s_i))}
\label{eq:Lc}
\end{equation}
where $\gamma$ is a parameter for weighting down the loss caused by easy samples, and it is set to $2$ according to \cite{lin2017focal}. Also, metric learning policies, like \cite{deng2019arcface}\cite{wang2018cosface}\cite{liu2017sphereface}, can be applied to train a better embedding function.


\subsection{Self-supervised learning for domain transfer}

According to the studies in \cite{chen2020simple}, minimizing the distances between the embeddings from random cropped and color distorted images can achieve the best result. However, applying color distortion may cause the racial bias, and we do not find obvious difference between using mirroring and random cropping in experiments. For each image, $x$, we only use it and its mirror, $x'$, for training, so a self-supervised learning loss called SimSiam \cite{chen2020exploring} can be expressed as:
\begin{equation}
L_s = \frac{1}{2}[D(h(z'), \varphi(z)) + D(h(z), \varphi(z'))]
\label{eq:Ls}
\end{equation}
\begin{equation}
D(p,z)=-\frac{p^{T}z}{\|p\|\|z\|}
\end{equation}
where $z=f(x)$, $z'=f(x')$, $\varphi$ is a stop-gradient operation, and $D(p,z)$ is a function for computing the cosine similarity between embeddings $p$ and $z$. Specially, $h$ is a head which remap the embeddings in another view, and its effectiveness has been discussed in \cite{chen2020simple} and \cite{chen2020exploring}.

To perform domain adaptation, we apply the loss on both source and target datasets and sum them up by a ratio which we call it adapting ratio, $\rho$. Thus, the loss is expressed as:
\begin{equation}
L_a = (1 - \rho)L_s(x^s) + {\rho}L_s(x^t) 
\label{eq:La}
\end{equation}
where $\rho$ controls the weights of SimSiam loss on two domains. For the purpose of domain transfer, it is reasonable to set the value of $\rho$ larger than $0.5$ a little bit. Since the loss only focus on self-similarity, it is necessary to use cross entropy loss to maintain the inter-class discrepancy. The total loss is defined as:
\begin{equation}
L = L_c + L_a
\end{equation}
We call it SimSiam Adapting (SSA) Loss, and use it to do the experiments in the next section.



\section{Experiments}

\input{table/ablation_study/adapt_ratio}
\input{table/ablation_study/embedding}
\input{table/benchmark_comparison/verification}
\input{table/benchmark_comparison/identification}

\subsection{Datasets}

In the experiments, CASIA-WebFace \cite{yi2014learning} is utilized to be the source dataset. It contains 10,575 identities and 494,414 images which are the pictures of celebrities on Internet. As for the testing datasets, we use IJB datasets, including IJB-A/B/C \cite{klare2015pushing}\cite{whitelam2017iarpa}\cite{maze2018iarpa}. IJB datasets contain mixture of images and videos in the wild. The conditions are all challenging due to blurry frames and large pose variations. IJB-A \cite{klare2015pushing} contains 500 identities with 5,396 images and 20,412 video frames. IJB-B \cite{whitelam2017iarpa} extended from IJB-A contains 1,845 identities with 11,755 images and 7,018 videos, and 10,044 non-face images. IJB-C \cite{maze2018iarpa} extended from IJB-B contains 3,531 identities with 21,294 images and 11,779 videos, and 10,040 non-face images. 

We follow the protocols in \cite{klare2015pushing} to evaluate the performance of verification, open-set and close-set identification, and there metrics are True Positive Rate vs. False Positive Rate (TPR@FAR), True Positive Identification Rate vs. set False Positive Identification Rate (TPIR@FAIR), and top-K accuracy (Rank-K) respectively.


\subsection{Implementation detail}

Cross entropy loss (\ref{eq:Lc}) is applied to train MobileFaceNet \cite{chen2018mobilefacenets} from scratch by CASIA-WebFace dataset \cite{yi2014learning}. The training epoch and batch size is set to 50 and 128 respectively. The learning rate is set to 0.1 and is divided by 10 every 12 epoch. The trained model is the baseline for domain adaptation, and its performance is the baseline in the experiments.

We treat all images in IJB-A dataset \cite{klare2015pushing} as the target dataset, so most images in IJB-B and IJB-C are not covered. As the unlabeled data from a target dataset are mixed in, the learning rate is decayed to be 0.0001 to protect the learned knowledge. The training epoch, batch size and learning rate schedule are preserved, but the epoch is counted based on the size of the target dataset, which induces less training iterations. For each batch, 128 images are sampled from each dataset. 

The head, $h$, used in SimSiam loss is a two-layer neural network, and the number of hidden neurons is set to 32 which is 4 times less than the embedding dimension of MobileFaceNet \cite{chen2018mobilefacenets}. According to \cite{chen2020exploring}, we add batch normalization and rectified linear unit in the hidden layer to guarantee its performance.


\subsection{Ablation study}
\label{ssec:ablation}

It is more efficient to use smaller dataset to evaluate the trained models in ablation study, so only IJB-A dataset \cite{klare2015pushing} is adopted in this part of experiments.

\subsubsection{Adapting ratio}

To find out the proper adapting ratio, $\rho$, we vary its value, and compare with the baseline. The comprehensive results are listed in Table \ref{tab:ablation_ratio}. It is obvious that the performances of verification at lower FPR and open-set identification are increased. Adopting SimSiam loss \cite{chen2020exploring} directly on source dataset only, $\rho=0.0$ can achieve some improvements, which shows its competency of generalization. Since larger weighting on target domain may limit the cluster learning relying on the supervision of source data and labels, overemphasis of learning self-similarity on target domain cannot preserve the inter-class discrepancy. As for verification at higher FPR and close-set identification, there is no improvement. Our hypothesis for this issue is discussed in the next part. According to the results in Table \ref{tab:ablation_ratio}, we choose $\rho=0.6$ as our best parameter setting.


\subsubsection{Embedding analysis}

How the embedding spaces are learned is discussed here. We compare the embedding distributions of the baseline, source-only case, $\rho=0.0$, and the best case, $\rho=0.6$. The averages of three similarities (mirror, intra-class, and inter-class) and embedding length are carried out in Table \ref{tab:ablation_embedding}.

Although the mirror similarities are increased with the guidance of SSA, there is no obvious changes on intra-class similarities, but they are a little bit lower. Oppositely, the inter-class similarities are reduced, and the embedding lengths are also enlarged, which implies the discrepancies among identities are increased. Since higher TPR or TPIR at lower FPR and FPIR requires lower inter-class similarity, SSA can do better under these protocols. Due to giant negative pairs in the protocols, the significant improvements can be achieved by little decay on inter-class similarity. On the other hand, higher TPR at higher FPR or Rank-K requires much higher intra-class similarity, so this is the reason why SSA fails under these protocols. 



\subsection{Benchmark comparison}

To guarantee the effectiveness of SSA, it is compared with the state-of-the-arts focusing on domain adaptation for face recognition. We also use a margin penalty method (ArcFace) proposed in \cite{deng2019arcface} to train our backbone to be another baseline. The comprehensive evaluations on IJB-A \cite{klare2015pushing}, IJB-B \cite{whitelam2017iarpa}, and IJB-C \cite{maze2018iarpa} are listed in Table \ref{tab:benchmark_ver} and Table \ref{tab:benchmark_id}. 

From Table \ref{tab:benchmark_ver}, we can observe that SSA can successfully improve the baselines on all benchmarks under almost all FPRs especially under lower FPRs. With the guidance of ArcFace \cite{deng2019arcface}, the performance can be better. However, in the identification protocols, Table \ref{tab:benchmark_id}, the improvements only exists on open-set protocols. Such issue has been discussed in \ref{ssec:ablation}. Since the models are adapted on IJB-A only \cite{klare2015pushing}, the performances of on the open-set protocols of IJB-B \cite{whitelam2017iarpa} and IJB-C \cite{maze2018iarpa} are not that good, but it can be further refined by adapting more data from these datasets. Expect for close-set identification protocols, compared with the state-of-the-arts, our approach shows its good performance by not only the improvements but also the much lighter backbone.



\section{Conclusion}

We focus on the unique problem of domain discrepancy in face recognition whose classes in domains are non-overlapping. Self-Supervised Adapting (SSA) loss is proposed in this paper. By adding an adapting ratio between the self-similarity losses on source and target domain, SSA can successfully improve the baseline models both verification and open-set identification protocols. Interestingly, we find that this progress is achieved by reducing inter-class similarities rather than increasing intra-class similarities through the analysis on the embedding distributions. Compared with other adapting methods under comprehensive protocols, SSA shows its competitive performance. However, it seems that SSA cannot preserve or even improve the intra-class similarity on target domain, so some advanced researches should be done in the future to compensate this problem.



\bibliographystyle{IEEEbib}
\bibliography{ref}

\end{document}

%% file: table/ablation_study/adapt_ratio.tex
\begin{table}[hb]
\caption{Evaluations on IJB-A \cite{klare2015pushing} with different $\rho$.}
\label{tab:ablation_ratio}
\centering
\begin{adjustbox}{max width=\linewidth}
\begin{tabular}{lcccccccc}  
\toprule
\multirow{2}{*}{Method} & \multicolumn{3}{c}{\bf Verification TPR (\%)} && \multicolumn{4}{c}{\bf Identification TPIR (\%)} \\
\cline{2-4}\cline{6-9} & FPR=0.001 & FPR=0.01 & FPR=0.1 && FPIR=0.01 & FPIR=0.1 & Rank-1 & Rank-10 \\
\hline
\hline
Baseline		& 75.63 & 90.54 & \textbf{96.88} && 65.31 & 85.74 & \textbf{94.79} & \textbf{97.91} \\
\hline
$\rho=0.0$		& 79.37 & 90.93 & 96.86 && 71.59 & 87.85 & 94.57 & 97.85 \\
$\rho=0.5$		& 80.78 & 91.03 & 96.70 && 73.23 & 87.82 & 94.74 & 97.78 \\
$\rho=0.6$		& \textbf{82.13} & \textbf{91.45} & 96.39 && \textbf{75.77} & \textbf{87.92} & 93.99 & 97.63 \\
$\rho=0.7$		& 79.57 & 90.86 & 96.28 && 70.11 & 86.71 & 94.13 & 97.63 \\
$\rho=0.8$		& 78.29 & 90.08 & 95.99 && 68.12 & 85.76 & 93.93 & 97.19 \\
$\rho=0.9$		& 74.33 & 90.12 & 96.06 && 64.36 & 84.91 & 93.72 & 97.24 \\
\bottomrule
\end{tabular}
\end{adjustbox}
\end{table}

%% file: table/ablation_study/embedding.tex
\begin{table}[hb]
\caption{Statistics of embedding metrics in IJB-A dataset \cite{klare2015pushing}.}
\label{tab:ablation_embedding}
\centering
\begin{adjustbox}{max width=\linewidth}
\begin{tabular}{lccccc}  
\toprule
\multirow{2}{*}{Method} & \multicolumn{3}{c}{\bf Similarity} && \multirow{2}{*}{\bf Embedding Length} \\
\cline{2-4} & Mirror & Intra-class & Inter-class && \\
\hline
\hline
Baseline			& 0.9478 & \textbf{0.7074} & 0.0728 && 113.89 \\
\hline
SSA ($\rho=0.0$)	& \textbf{0.9583} & 0.6915 & 0.0199 && \textbf{119.68} \\
SSA ($\rho=0.6$)	& 0.9547 & 0.6905 & \textbf{0.0166} && 119.55 \\
\bottomrule
\end{tabular}
\end{adjustbox} 
\end{table}

%% file: table/benchmark_comparison/verification.tex
\begin{table*}[t]
\begin{minipage}{\textwidth}
\caption{Verification performance on IJB-A \cite{klare2015pushing}, IJB-B \cite{whitelam2017iarpa}, and IJB-C \cite{maze2018iarpa}. The bold texts stand for the highest TPIR in a column. The text with underline means it is better than baseline.}
\label{tab:benchmark_ver}
\centering
\begin{adjustbox}{max width=\linewidth}
\begin{tabular}{lcccccccccccccc}  
\toprule
\multirow{2}{*}{Method} & \multicolumn{4}{c}{\bf IJB-A TPR (\%)} && \multicolumn{4}{c}{\bf IJB-B TPR (\%)} && \multicolumn{4}{c}{\bf IJB-C TPR (\%)} \\
\cline{2-5}\cline{7-10}\cline{12-15} & FPR=0.0001 & FPR=0.001 & FPR=0.01 & FPR=0.1 && FPR=0.0001 & FPR=0.001 & FPR=0.01 & FPR=0.1 && FPR=0.0001 & FPR=0.001 & FPR=0.01 & FPR=0.1 \\
\hline
\hline
Sohn et al. \cite{sohn2017unsupervised}		& - & 58.40 & 82.80 & 96.20 && - & - & - & - && - & - & - & - \\
IMAN-A \cite{wang2019racial}				& - & 84.49 & 91.88 & 97.05 && - & - & - & - && - & - & - & - \\
CDA(vgg-soft) \cite{wang2020deep}			& - & 74.76 & 89.76 & \textbf{98.19} && - & - & - & - && - & - & - & - \\
CDA(res-arc) \cite{wang2020deep}			& - & 82.45 & 91.11 & 96.96 && - & 87.35 & 94.55 & 98.08 && - & 88.06 & 94.85 & 98.33 \\
\hline
SoftMax\footnote{This is baseline model trained by cross entropy on source dataset only.}										
											& 52.23 & 75.63 & 90.54 & 96.88 && 68.91 & 83.61 & 93.43 & 98.22 && 74.04 & 86.44 & 94.59 & 98.54 \\
ArcFace\footnote{This is another baseline model trained by the guidance of margin penalty proposed in \cite{deng2019arcface}.} \cite{deng2019arcface}
											& 72.60 & 84.82 & 92.18 & 96.11 && 77.29 & 87.18 & 94.25 & 98.26 && 81.33 & 89.75 & 95.35 & 98.48 \\
\textbf{SSA-SoftMax (ours)}					& \underline{62.51} & \underline{82.13} & \underline{91.45} & 96.39 && \underline{71.22} & \underline{84.88} & \underline{93.78} & 98.21 && \underline{75.61} & \underline{87.47} & \underline{94.92} & 98.47 \\
\textbf{SSA-ArcFace (ours)}					& \underline{\textbf{78.18}} & \underline{\textbf{87.37}} & \underline{\textbf{92.41}} & 95.84 && \underline{\textbf{78.48}} & \underline{\textbf{88.27}} & \underline{\textbf{94.88}} & \underline{\textbf{98.49}} && \underline{\textbf{82.72}} & \underline{\textbf{90.91}} & \underline{\textbf{95.90}} & \underline{\textbf{98.62}} \\
\bottomrule
\end{tabular}
\end{adjustbox} 
\end{minipage}
\end{table*}

%% file: table/benchmark_comparison/identification.tex
\begin{table*}[t]
\begin{minipage}{\textwidth}
\caption{Identification performance on IJB-A \cite{klare2015pushing}, IJB-B \cite{whitelam2017iarpa}, and IJB-C \cite{maze2018iarpa}. The bold texts stand for the highest TPIR in a column. The text with underline means it is better than baseline.}
\label{tab:benchmark_id}
\centering
\begin{adjustbox}{max width=\linewidth}
\begin{tabular}{lcccccccccccccc}  
\toprule
\multirow{2}{*}{Method} & \multicolumn{4}{c}{\bf IJB-A TPIR (\%)} && \multicolumn{4}{c}{\bf IJB-B TPIR (\%)} && \multicolumn{4}{c}{\bf IJB-C TPIR (\%)} \\
\cline{2-5}\cline{7-10}\cline{12-15} & FPIR=0.01 & FPIR=0.1 & Rank-1 & Rank-10 && FPIR=0.01 & FPIR=0.1 & Rank-1 & Rank-10 && FPIR=0.01 & FPIR=0.1 & Rank-1 & Rank-10\\
\hline
\hline
Sohn et al. \cite{sohn2017unsupervised}		& - & - & 87.90 & 97.00 && - & - & - & - && - & - & - & - \\
IMAN-A \cite{wang2019racial}				& - & - & 94.05 & 98.04 && - & - & - & - && - & - & - & - \\
CDA(vgg-soft) \cite{wang2020deep}			& 66.85 & 85.32 & \textbf{94.89} & \textbf{99.23} && - & - & - & - && - & - & - & - \\
CDA(res-arc) \cite{wang2020deep}			& 75.49 & 87.76 & 93.61 & 97.62 && - & - & 86.22 & 93.33 && - & - & 88.19 & 93.70 \\
\hline
SoftMax\footnote{This is baseline model trained by cross entropy on source dataset only.}										
											& 65.31 & 85.74 & 94.79 & 97.91 && 59.77 & 77.10 & 88.01 & 95.22 && 58.86 & 76.49 & 88.93 & 95.07 \\
ArcFace\footnote{This is another baseline model trained by the guidance of margin penalty proposed in \cite{deng2019arcface}.} \cite{deng2019arcface}
											& 78.99 & 88.75 & 94.58 & 97.50 && \textbf{66.22} & 81.23 & \textbf{89.49} & \textbf{95.37} && \textbf{70.49} & 81.87 & \textbf{90.61} & \textbf{95.60} \\
\textbf{SSA-SoftMax (ours)}					& \underline{75.77} & \underline{87.92} & 93.99 & 97.63 && 58.74 & \underline{77.51} & 86.98 & 94.41 && 57.24 & \underline{77.69} & 87.92 & 94.50 \\
\textbf{SSA-ArcFace (ours)}					& \underline{\textbf{80.03}} & \underline{\textbf{89.47}} & 94.26 & 97.40 && 64.47 & \underline{\textbf{82.48}} & 89.31 & 95.23 && 69.80 & \underline{\textbf{83.68}} & 90.50 & 95.47 \\
\bottomrule
\end{tabular}
\end{adjustbox} 
\end{minipage}
\end{table*}